\documentclass[11pt]{article}

\usepackage[preprint]{acl}

\usepackage{times}
\usepackage{latexsym}

\usepackage[T1]{fontenc}

\usepackage[utf8]{inputenc}

\usepackage{microtype}

\usepackage{inconsolata}

\usepackage{graphicx}

\usepackage{makecell}
\usepackage{amsmath}
\usepackage{amsfonts}
\usepackage{algorithm}
\usepackage{algorithmic}
\usepackage{amssymb} 
\usepackage{graphicx}
\usepackage{subcaption} 
\usepackage{booktabs}
\usepackage{multirow}
\usepackage{tcolorbox}
\usepackage{listings}
\usepackage{enumitem}
\newtcolorbox{promptbox}{
  colback=gray!5,
  colframe=gray!60,
  boxrule=0.6pt,
  arc=2pt,
  left=6pt,
  right=6pt,
  top=6pt,
  bottom=6pt
}
\newcommand{\ie}{\emph{i.e., }}

\newcommand{\cf}{\emph{cf. }}

\title{Confidence Before Answering: A Paradigm Shift for Efficient LLM Uncertainty Estimation}

\author{
 \textbf{Changcheng Li\textsuperscript{1}},
 \textbf{Jiancan Wu\textsuperscript{1}\thanks{Jiancan Wu and Qi Tian are the corresponding authors}},
 \textbf{Hengheng Zhang\textsuperscript{2}},
 \textbf{Zhengsu Chen\textsuperscript{2}},
 \textbf{Guo An\textsuperscript{2}}, \\
 \textbf{Junxiang Qiu\textsuperscript{1}},
 \textbf{Xiang Wang\textsuperscript{1}},
 \textbf{Qi Tian\textsuperscript{2}\textsuperscript{*}}
\\  
 \textsuperscript{1}University of Science and Technology of China,
 \textsuperscript{2}Huawei Inc.
\\  
 \small  
   \texttt{\{lichangcheng, qiujx\}@mail.ustc.edu.cn, 
   \{wujcan, xiangwang1223\}@gmail.com} \\
   \small
   \texttt{\{zhanghengheng55, chenzhengsu2, anguo1, tian.qi1\}@huawei.com}
}

\begin{document}

\maketitle

\begin{abstract}
Reliable deployment of large language models (LLMs) requires accurate uncertainty estimation. Existing methods are predominantly answer-first, producing confidence only after generating an answer, which measure the correctness of a specific response and limits practical usability. 
We study a confidence-first paradigm, where the model outputs its confidence before answering, interpreting this score as the model’s probability of answering the question correctly under its current policy.

We propose \textbf{CoCA(Co-optimized Confidence and Answers)}, a GRPO reinforcement learning framework that jointly optimizes confidence calibration and answer accuracy via segmented credit assignment. By assigning separate rewards and group-relative advantages to confidence and answer segments, CoCA enables stable joint optimization and avoids reward hacking. Experiments across math, code, and factual QA benchmarks show improved calibration and uncertainty discrimination while preserving answer quality, thereby enabling a broader range of downstream applications.
\end{abstract}
\section{Introduction}

\begin{figure*}[t]
    \centering
    \includegraphics[
    width=0.98\linewidth,
    trim={0cm 2.4cm 0cm 0.2cm},
    clip
    ]{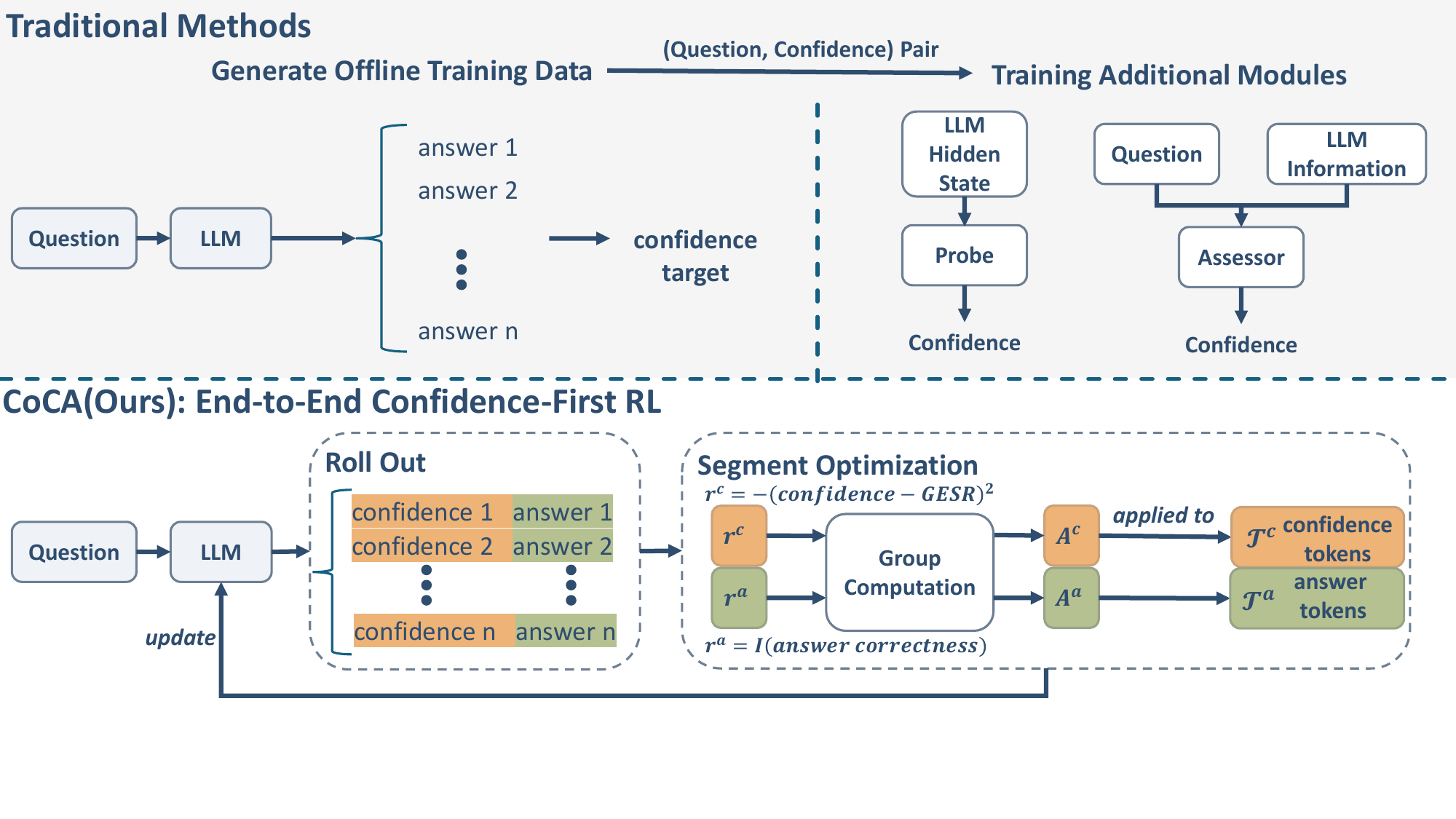} 
    \caption{
    \textbf{From Decoupled Confidence Estimation to End-to-End Confidence-First Learning.}
\textit{Top}: Traditional pipelines derive confidence targets from offline model outputs and train separate predictors on frozen correctness targets.
\textit{Bottom}: CoCA (ours) jointly generates confidence and answers and optimizes them end-to-end with segment-specific GRPO rewards, using group-wise empirical success rates (GESR) as dynamic confidence targets.
    }
    \label{fig:confans-grpo-overview}
\end{figure*}

LLMs have made remarkable progress on reasoning-intensive tasks, yet hallucinations remain pervasive --- they frequently generate plausible but incorrect responses \cite{Ji2023,Bang2023}.
This problem may be amplified by current post-training paradigms \cite{mei2025,kirichenko2025}, resulting in overconfidence that undermines trustworthiness, particularly in high-stakes domains such as medicine \cite{Pal2023}, law \cite{dahl2024}, and finance \cite{joshi2025}.
Recognizing this challenge, a growing body of work has studied \textit{confidence estimation} in LLMs \cite{kadavath2022a,stangel2025} --- methods that produce a numerical score reflecting
how likely the model’s answer is to be correct.
Well-calibrated confidence estimates not only help users judge answer reliability, but also support system-level decisions such as selective answering, refusal, and model routing \cite{2025Chen}.

Most existing methods estimate confidence in an \textit{answer-first} manner, which generates responses before estimating confidence through internal probing \cite{mielke2022,fadeeva2024}, post-hoc verbalized confidence \cite{lin2022,xu2024}, or sampling-based surrogates \cite{aichberger2024}.
They essentially ask ``\textit{Is the specific answer correct?}'', but incur high computational overhead and cannot enable early decisions.
In contrast, \textbf{confidence-first} approaches estimate correctness probability before generation, asking a fundamentally harder question --- ``\textit{Given my current capabilities, how likely am I to answer correctly?}''.
Toward this goal, existing methods typically train separate supervised modules on frozen correctness labels.
They generate the LLM's answers on the training dataset, label each by correctness, then train a confidence predictor --- either on the model's internal representations \cite{Cencerrado2025} or an external accessor \cite{Zhou2022} --- to predict these frozen labels.

Despite effectiveness, this decoupled pipeline faces two fundamental challenges:
\begin{itemize}[leftmargin=*]
    \item \textbf{Confidence estimation is inherently policy-dependent}.
    Training on frozen correctness labels usually causes predictors to overfit to superficial patterns (such as problem difficulty), rather than capturing the model's intrinsic uncertainty \cite{Farquhar2024}.
    Proper confidence optimization therefore requires tracking the dynamic evolution of the model’s capability to prevent such optimization hacking.
\item \textbf{Confidence and answer quality are intrinsically entangled}.
Users care about both reliable confidence estimates and accurate answers.
However, isolated
confidence training can degrade answer quality\cite{damani2025a}. Jointly optimizing confidence and accuracy can alleviate this mismatch, but it demands precise credit assignment to enable stable end-to-end learning \cite{ha2025,guo2025}, since confidence tokens and answer tokens are governed by distinct optimization objectives.
\end{itemize}

To address these challenges, we propose \textbf{CoCA (Co-optimized Confidence and Answers)}, an end-to-end, confidence-first learning framework that jointly optimizes confidence calibration and answer quality without requiring separate modules or frozen labels.
The key idea is to have the model verbalize its confidence before generating the answer, then co-optimize both through a unified policy gradient objective with segment-specific credit assignment.
Specifically, we build upon Group Relative Policy Optimization (GRPO) \cite{guo2025deepseek}, and introduce three core designs:
(1) \textbf{Dynamic confidence targets}. Rather than training on static correctness labels, we align confidence targets to group-wise empirical success rates (GESR) observed during policy rollouts. By grounding confidence in the model's real-time performance, estimates naturally track evolving model capabilities without requiring re-labeling.
(2) \textbf{Calibration rewards}. We incorporate a Brier score penalty \cite{BRIER1950} (\ie the squared difference between expressed confidence and GESR, \cf Equation \eqref{eq:brier}) into the reward function to quantify miscalibration. This quadratic form amplifies penalties for severe miscalibration --- confident but wrong predictions or hesitant but correct ones --- thereby incentivizing the model to accurately reflect its capability.
(3) \textbf{Segment-specific reward decomposition}. Each response receives targeted rewards for its two segments: the confidence segment is rewarded for calibration accuracy, while the answer segment is rewarded for task correctness, preventing the model from sacrificing answer quality to improve calibration during optimization.

Experiments show that when trained only on math datasets, CoCA attains strong calibration not only in-distribution but also under distribution shift --- for example, on Qwen2.5-3B-Instruct it reduces ECE from 0.54 to 0.09 on Math and from 0.66 to 0.14 on Factual QA, outperforming existing confidence-first baselines. Moreover, compared to answer-first methods, it enables much earlier decision-making by emitting confidence with only \textasciitilde10 tokens, and cutting confidence-estimation token cost by >92\% across all categories.

\section{Related Work}
\subsection{Answer-first Confidence Estimation}

\textbf{Internal Probing.} A common approach is to probe a model’s internal states or output probabilities to estimate confidence in a given answer. \citet{kadavath2022a} prompt language models to output ``true'' or ``false'' and use the probability of ``true'' as a proxy for confidence. \citet{mielke2022} condition response generation on external confidence probes. \citet{fadeeva2024} propose Claim Conditioned Probability, a token-level uncertainty method based on internal signals. \citet{azaria2023,orgad2024} show that hidden states encode truthfulness cues, while \citet{kapoor2025} introduce an auxiliary uncertainty head fine-tuned via LoRA. 

\textbf{Post-hoc Verbalized Confidence.} Another line of work elicits verbalized confidence (numeric or natural-language\citep{tao2025,zhang2023RTuning}) from the LLM after answering, and calibrates the resulting confidence behavior using supervised fine-tuning or reinforcement learning.\citet{lin2022} train GPT-3 to estimate confidence directly by regressing on its empirical accuracy over question–answer pairs. \citet{stengel-eskin2024} propose a speaker–listener setup where the speaker is rewarded based on the listener’s inferred confidence. \citet{leng2024} integrate explicit confidence annotations into reward model training, improving alignment with verbalized confidence levels. \citet{xu2024} and \citet{stangel2025} apply reinforcement learning with proper scoring rules as rewards --- using the Brier score and a clipped log loss, respectively to enhance calibration. In contrast, \citet{damani2025a} use a single reward to jointly optimize confidence and accuracy.

\textbf{Sampling-based Surrogates.} This line of work leverages response agreement, such as majority voting or best-of-$N$ sampling, as a proxy for confidence. \citet{aichberger2024} generate semantically diverse yet plausible outputs and assess uncertainty via their consistency. \citet{kuhn2022} introduce semantic entropy, a sampling-based method that accounts for linguistic variations to better capture uncertainty in natural language generation. \citet{xue2025} assess model uncertainty by introducing cross-model consistency. 

\subsection{Confidence-first Confidence Estimation}

In contrast to the extensive body of work on confidence estimation for specific answers, this area remains relatively underexplored. A number of studies investigate whether a model is able to answer a question by probing its internal representations \cite{Ferrando2025,Cencerrado2025,2025Chen, kadavath2022a}. Specifically, \citet{Ferrando2025} decompose intermediate model layers (the residual stream) using Sparse Autoencoders (SAEs) to determine whether the model recognizes a given entity, while \citet{Cencerrado2025} employ probes to predict confidence for a given question. Other works rely on external assessors for evaluation, where the assessors range from neural networks \cite{Orallo2022} to Random Forests \cite{Zhou2022}, as well as XGBoost and Logistic Regression models \cite{Pacchiardi2025}. Beyond these approaches, \citet{Shrivastava2025} obtain confidence by asking the model to perform pairwise comparisons across questions and ranking them accordingly.

Given the different application scopes, target signals, and evaluation protocols of these studies, we compare against representative confidence-first mechanisms, such as internal probes and external assessors, rather than reproducing every individual system.
\section{Method}
\subsection{Preliminaries: RL for LLMs and GRPO}

Given an input prompt $x$, we denote the language model policy as $\pi_\theta(\cdot\mid x)$, which generates a token sequence $y=(y_1,\dots,y_T)$. In reinforcement learning for LLMs (e.g., RLHF/RLAIF/RLVR) \citep{bai2022,bai2022constitutional,lee2023rlaif,guo2025deepseek}, the standard objective is to maximize an external reward $R(x,y)$ while preventing the policy from drifting too far from a reference policy $\pi_{\text{ref}}$.

\textbf{GRPO (Group Relative Policy Optimization)} \citep{guo2025deepseek} is a PPO-style method that avoids training an explicit value function. For each prompt $x$, GRPO samples a group of $G$ candidate responses from the current policy, computes a scalar reward $r_i$ for each response, and constructs a group-wise relative advantage to reduce variance:
\begin{equation}
    \hat{A}_i = \frac{r_i - \mu(r)}{\sigma(r)+\epsilon},
\end{equation}

\noindent where $\mu(r)$ and $\sigma(r)$ are the mean and standard deviation computed over the $G$ rewards. Let $\pi_{\theta_{\text{old}}}$ be the policy before the update. Define the token-level probability ratio as follows:
\begin{equation}
\rho_{i,t}(\theta)=\frac{\pi_\theta(y_{i,t}\mid x, y_{i,<t})}{\pi_{\theta_{\text{old}}}(y_{i,t}\mid x, y_{i,<t})}.
\end{equation}

Then a clipped GRPO objective can be written as:
\begin{equation}
\begin{aligned}
\mathcal{L}_{\text{GRPO}}(\theta)=&
\mathbb{E}_{x}\Bigg[\frac{1}{G}\sum_{i=1}^G \sum_{t=1}^{T_i}
\min\big(\rho_{i,t}(\theta)\hat{A}_i, \\ & \mathrm{clip}(\rho_{i,t}(\theta),1-\varepsilon,1+\varepsilon)\hat{A}_i\big)\Bigg]
\\ &-\ \beta\mathbb{E}_{x}\big[\mathrm{KL}(\pi_\theta||\pi_{\text{ref}})\big].
\end{aligned}
\end{equation}
The formulation above uses a \textbf{single reward} to drive the \textbf{entire} response sequence.

\subsection{Confidence-First Paradigm Definition}

We study a confidence-first paradigm: the model must output its confidence before producing the answer. We decompose the output into two segments:
\begin{equation}
    y = (y^{c}, y^{a}),
\end{equation}
where $y^{c}$ is the confidence segment and $y^{a}$ is the answer segment. We enforce a fixed format:
\begin{equation}
y \equiv \texttt{<confidence>} \ s \ \texttt{</confidence>} \  y^{a}.
\end{equation}

Training a confidence-first model is inherently a \textbf{multi-objective problem}: the policy must output a calibrated confidence score and produce a correct answer. 

\subsection{CoCA: Segmented GRPO for Co-optimized Confidence and Answers}
\subsubsection{Reward Formulation}

\textbf{Accuracy reward}. For each prompt $x$, we sample $G$ full outputs $y_i=(y_i^c,y_i^a)$. We define an answer correctness reward ($r^{a}_{i} \in \{0,1\}$) as

\begin{equation}
    r^{a}_i = \mathbb{I}\big(\text{AnsCorrect}(x, y_i^a)\big),
\end{equation}

where $\text{AnsCorrect}(\cdot)$ is computed by the dataset-specific evaluator.

\noindent\textbf{Confidence Reward.} Next, we define GESR as an estimate of how likely the current policy answers this question correctly:

\begin{equation}
\hat{p}(x) = \frac{1}{G}\sum_{j=1}^G r^{a}_j.
\end{equation}
The confidence segment is parsed into a scalar $s_i = \text{Parse}(y_i^c)\in[0,1]$. We encourage $s_i$ to match $\hat{p}(x)$ using a stable \textbf{Brier-style reward}:

\begin{equation} \label{eq:brier} 
r^{c}_i = -\big(s_i - \hat{p}(x)\big)^2.
\end{equation}
Throughout the entire process, the confidence target is derived from the same rollout via the GESR $\hat{p}(x)$. Meanwhile, we do not employ any sampling strategy and instead preserve the model’s original distribution. This makes $s$ reflect the probability of answering correctly under the current policy.

\subsubsection{Segmented Credit Assignment and Joint Optimization}

Sequentially optimizing accuracy and then confidence can introduce reward hacking: the model may improve the confidence objective by altering answer behavior (e.g., refusal or evasiveness). CoCA avoids this by optimizing both objectives simultaneously, while restricting each advantage to its corresponding token span, which anchors answer quality and confidence calibration throughout training.

We therefore compute two advantages within the same group:

\begin{equation}
\hat{A}^{c}_i
=\frac{r^{c}_i-\mu(r^{c})}{\sigma(r^{c})+\epsilon},
\qquad
\hat{A}^{a}_i
=\frac{r^{a}_i-\mu(r^{a})}{\sigma(r^{a})+\epsilon}.
\end{equation}

We then apply the clipped policy gradient separately to the confidence and answer token segments. Let $\mathcal{T}^c_i$ denote the set of tokens in the confidence segment of sample $i$, and $\mathcal{T}^a_i$ denote those in the answer segment. Our segmented objective, without a KL-divergence term, is given by

\begin{equation}
\begin{aligned}
\mathcal{L}^{c}_i(\theta)
&= \sum_{t\in \mathcal{T}^c_i}
\min\Big(
\rho_{i,t}(\theta)\hat{A}^{c}_i, \\
&\qquad
\mathrm{clip}(\rho_{i,t}(\theta),1-\varepsilon,1+\varepsilon)\hat{A}^{c}_i
\Big), \\
\mathcal{L}^{a}_i(\theta)
&= \sum_{t\in \mathcal{T}^a_i}
\min\Big(
\rho_{i,t}(\theta)\hat{A}^{a}_i, \\
&\qquad
\mathrm{clip}(\rho_{i,t}(\theta),1-\varepsilon,1+\varepsilon)\hat{A}^{a}_i
\Big).
\end{aligned}
\end{equation}
Here $\rho_{i,t}(\theta)$ and $\mathrm{clip}(\cdot)$ follow standard PPO/GRPO definitions. By segmenting the output $y=(y^c,y^a)$ and computing separate advantages $\hat{A}^c$ and $\hat{A}^a$ that are applied only to their respective token spans, CoCA provides a more targeted learning signal and leads to faster and more stable training.

The \textbf{joint optimization} is as follows: 
\begin{equation}
\begin{aligned}
\mathcal{L}_{\text{CoCA}}(\theta)
&= \mathbb{E}_{x}\Bigg[
\frac{1}{G}\sum_{i=1}^G
\Big(
\mathcal{L}^{c}_i(\theta)
+ \mathcal{L}^{a}_i(\theta)
\Big)
\Bigg].
\end{aligned}
\end{equation}

The complete algorithmic workflow is presented in Algorithm~\ref{alg:coca}.

\section{Experiment}
This section primarily examines whether, under the confidence-first paradigm, CoCA can improve the usability and cross-domain generalization of confidence estimates while preserving answer quality. 
In addition, we perform comparisons against the answer-first paradigm to assess whether CoCA can attain comparable performance while reducing the token cost of obtaining confidence estimates.
We also conduct ablation studies to contrast segmented versus joint rewards, and to expose reward hacking risks arising from sequential training.

\subsection{Experimental Setup}
\subsubsection{Models and Training Data}
We conduct our confidence-first comparisons on three instruction-tuned \textbf{Qwen2.5-Instruct} models of different scales: 7B, 3B, and 1.5B~\cite{qwen2.5}, to verify consistency across model sizes.
Unless otherwise specified, all remaining experiments (including answer-first comparisons and ablations) are conducted on \textbf{Qwen2.5-7B-Instruct}.

For the main experiments, training is performed on \textbf{Big-Math-Verified}~\citep{albalak2025}, a math dataset with automatically verifiable correctness, enabling low-noise reward computation.

\subsubsection{Evaluation Benchmarks}
After training, all models are evaluated on a diverse set of benchmarks:
\begin{itemize}[leftmargin=*]
    \item \textbf{Math}: AIME2024, AIME2025, MATH-500 \citep{hendrycks2021a}, GSM8K \citep{cobbe2021}
    \item \textbf{Code}: HumanEval \citep{chen2021}, Sanitized MBPP \citep{austin2021}
    \item \textbf{Factual QA}: SimpleQA \citep{wei2024}, TriviaQA \citep{joshi2017}
\end{itemize}

Importantly, although training is performed solely on math data, we evaluate on code and factual QA to test whether the learned confidence reflects general \textbf{uncertainty awareness} rather than domain-specific heuristics.

\subsubsection{Metrics}

\begin{itemize}[leftmargin=*]
    \item \textbf{Accuracy $\uparrow$}: The proportion of correct predictions among all samples.
    
    \item \textbf{AUROC $\uparrow$}: Measures how well confidence scores distinguish correct answers from errors.
    
    \item \textbf{Expected Calibration Error (ECE) $\downarrow$}: Measures the gap between predicted confidence and actual accuracy across different confidence bins.

    \item \textbf{Brier Score $\downarrow$}: Measures the mean squared error between confidence and binary correctness.

\end{itemize}
Additionally, in Section~\ref{sec:Answer-First}, we also measured the token consumption to confidence prediction (TTC), so as to reflect both computational cost and latency.

\subsection{Comparison with Confidence-First Baselines}

We first compare CoCA with approaches that either directly predict confidence or attach confidence estimation to an accuracy-optimized model.

\subsubsection{Baselines}
We consider the following baselines:
\begin{itemize}[leftmargin=*]
    \item \textbf{Verbalized Confidence}: eliciting the model's self-reported confidence by designing prompts that explicitly ask it to express its confidence level without confidence-specific optimization.
    \begin{itemize}[leftmargin=1.5em]
      \item \textbf{Instruct Model}: the original instruction-tuned model.
      \item \textbf{RLVR (Accuracy-only)}: reinforcement learning that optimizes only answer accuracy.
    \end{itemize}

    \item \textbf{Internal Probing}: extracting uncertainty-related signals from the model's internal states.
    \begin{itemize}[leftmargin=1.5em]
      \item \textbf{Question Probability}: using the likelihood of the question tokens as a confidence proxy.
      \begin{equation}
          \mathrm{QuestionProb}(x)\;=\;\frac{1}{|\mathcal{X}|}\sum_{i \in \mathcal{X}}P_\theta\!\left(x_i \mid x_{<i}\right).
      \end{equation}
      Here, $\mathcal{X}$ denotes the sequence of input tokens in the question, and $P_\theta(x_i \mid x_{<i})$ represents the model's probability of generating token $x_i$ conditioned on all preceding input tokens.
      \item \textbf{Probe}: a two-layer MLP probe trained on frozen hidden states to predict confidence.
    \end{itemize} 

    \item \textbf{Assessor Model}: estimating the model's confidence using an external model or algorithm.
    \begin{itemize}[leftmargin=1.5em]
      \item \textbf{Additional Assessor Model}: using Qwen2.5-1.5B-Instruct trained to predict the target model's correctness probability from the question alone.
    \end{itemize}
\end{itemize}

These baselines cover verbalized confidence methods, probing-based methods, likelihood-based heuristics, and external assessor models.

\subsubsection{Results and Analysis}

\begin{table*}[t]
\centering
\small
\setlength{\tabcolsep}{5pt}
\caption{\textbf{Main results compared with confidence-first methods}. ``Math/Code/Factual'' are benchmark-category averages. Bold indicates the best method per model $\times$ category and metric. For 1.5B, we also report confidence-generation success rate (SR) in Table~\ref{tab:appendix_qwen25_1p5b}.}

\begin{tabular}{c@{\hspace{3pt}}clccc ccc ccc}
\toprule
& & & \multicolumn{9}{c}{\textbf{Qwen2.5-Instruct (temp=1.0, pass@1)}} \\
\cmidrule(lr){4-12}
\textbf{Answer} & \textbf{Confidence} & \textbf{Metric}
& \multicolumn{3}{c}{\textbf{1.5B}}
& \multicolumn{3}{c}{\textbf{3B}}
& \multicolumn{3}{c}{\textbf{7B}} \\
\cmidrule(lr){4-6} \cmidrule(lr){7-9} \cmidrule(lr){10-12}
& & 
& \multirow{2}{*}{\textbf{Math}} & \multicolumn{2}{c}{\textbf{OOD}}
& \multirow{2}{*}{\textbf{Math}} & \multicolumn{2}{c}{\textbf{OOD}}
& \multirow{2}{*}{\textbf{Math}} & \multicolumn{2}{c}{\textbf{OOD}} \\
\cmidrule(lr){5-6} \cmidrule(lr){8-9} \cmidrule(lr){11-12}
& &
& & \textbf{Code} & \textbf{Factual}
& & \textbf{Code} & \textbf{Factual}
& & \textbf{Code} & \textbf{Factual} \\
\midrule

\multirow{4}{*}{Instruct}
& \multirow{4}{*}{Verbalized}
& Acc
& 13.79 & 55.10 & 14.81
& 37.54 & 63.78 & 22.71
& 43.73 & 77.43 & 30.13 \\
&& AUROC
& 0.52 & 0.53 & 0.53
& 0.54 & 0.52 & 0.57
& 0.61 & 0.57 & 0.63 \\
&& ECE
& 0.77 & 0.41 & 0.72
& 0.54 & 0.30 & 0.66
& 0.52 & 0.22 & 0.58 \\
&& Brier
& 0.73 & 0.42 & 0.66
& 0.52 & 0.33 & 0.60
& 0.50 & 0.23 & 0.52 \\
\midrule

\multirow{4}{*}{RLVR}
& \multirow{4}{*}{Verbalized}
& Acc
& 30.89 & 39.00 & 16.77
& 42.86 & 34.58 & 23.92
& 51.50 & 76.10 & 30.74 \\
&& AUROC
& 0.52 & 0.58 & 0.54
& 0.34 & 0.48 & 0.54
& 0.63 & 0.55 & 0.60 \\
&& ECE
& 0.62 & 0.56 & 0.76
& 0.48 & 0.56 & 0.62
& 0.46 & 0.23 & 0.58 \\
&& Brier
& 0.61 & 0.55 & 0.71
& 0.47 & 0.54 & 0.56
& 0.45 & 0.23 & 0.53 \\
\midrule

\multirow{4}{*}{RLVR}
& \multirow{4}{*}{QuestionProb}
& Acc
& 39.40 & 57.15 & 19.79
& 46.53 & 69.03 & 26.01
& 50.23 & 76.58 & 31.38 \\
&& AUROC
& 0.44 & 0.46 & 0.51
& 0.42 & 0.42 & 0.50
& 0.32 & 0.47 & 0.55 \\
&& ECE
& 0.36 & 0.48 & 0.19
& 0.44 & 0.65 & 0.26
& 0.47 & 0.70 & 0.30 \\
&& Brier
& 0.37 & 0.48 & 0.20
& 0.45 & 0.63 & 0.26
& 0.47 & 0.68 & 0.31 \\
\midrule

\multirow{4}{*}{RLVR}
& \multirow{4}{*}{Assessor}
& Acc
& 39.40 & 57.15 & 19.79
& 46.53 & 69.03 & 26.01
& 50.23 & 76.58 & 31.38 \\
&& AUROC
& 0.72 & 0.63 & 0.63
& 0.83 & 0.64 & 0.61
& 0.83 & 0.55 & 0.61 \\
&& ECE
& 0.16 & 0.25 & 0.16
& 0.16 & 0.29 & 0.27
& 0.24 & 0.29 & 0.26 \\
&& Brier
& 0.14 & 0.35 & 0.17
& 0.14 & 0.33 & 0.25
& 0.19 & 0.30 & 0.26 \\
\midrule

\multirow{4}{*}{RLVR}
& \multirow{4}{*}{Probe}
& Acc
& 39.40 & 57.15 & 19.79
& 46.53 & 69.03 & 26.01
& 50.23 & 76.58 & 31.38 \\
&& AUROC
& 0.62 & 0.64 & 0.57
& 0.84 & 0.59 & 0.64
& \textbf{0.84} & 0.62 & 0.64 \\
&& ECE
& 0.10 & 0.27 & 0.46
& 0.11 & 0.37 & 0.29
& 0.12 & 0.71 & \textbf{0.21} \\
&& Brier
& 0.12 & 0.34 & 0.42
& 0.11 & 0.39 & 0.24
& 0.10 & 0.68 & \textbf{0.21} \\
\midrule

\multicolumn{2}{c}{\multirow{4}{*}{\makecell{\textbf{CoCA(ours)}\\Joint optimization}}}
& Acc
& 36.85 & 46.72 & 18.89
& 44.85 & 67.38 & 24.74
& 47.92 & 76.60 & 28.85 \\
&& AUROC
& \textbf{0.82} & \textbf{0.72} & \textbf{0.71}
& \textbf{0.88} & \textbf{0.67} & \textbf{0.73}
& 0.71 & \textbf{0.72} & \textbf{0.69} \\
&& ECE
& \textbf{0.09} & \textbf{0.09} & \textbf{0.09}
& \textbf{0.09} & \textbf{0.28} & \textbf{0.14}
& \textbf{0.10} & \textbf{0.16} & 0.26 \\
&& Brier
& \textbf{0.10} & \textbf{0.22} & \textbf{0.12}
& \textbf{0.09} & \textbf{0.30} & \textbf{0.14}
& \textbf{0.09} & \textbf{0.19} & 0.24 \\

\bottomrule
\end{tabular}
\label{tab:main}
\end{table*}

Category-wise statistics are reported in Table~\ref{tab:main}, and detailed benchmark-level results are provided in Tables~\ref{tab:appendix_qwen25_7b}, \ref{tab:appendix_qwen25_3b}, and \ref{tab:appendix_qwen25_1p5b}.
Across all benchmarks, we observe the following consistent trends:

\textbf{Uncalibrated verbalized confidence is severely overconfident.}
Without explicit confidence calibration, both instruction-tuned models and accuracy-only RLVR models exhibit severe overconfidence in their verbalized confidence. For example, the 1.5B Instruct model achieves only 13.79\% accuracy on Math, but its ECE and Brier Score reach 0.77 and 0.73, respectively. These results show that improving answer correctness alone does not teach the model to express reliable uncertainty.

\textbf{External assessor models and probes are sensitive to distribution shift.}
The outputs of external assessor models tend to concentrate around 0.5, exhibiting limited discriminative power. While probes or auxiliary assessor models can improve in-domain calibration, these gains are not consistently preserved on out-of-domain code and factual QA tasks, indicating a dependence on the training distribution and learned representations.

\textbf{Question probability is a weak proxy for correctness.}
Question probability tends to assign uniformly low scores, and token likelihood primarily reflects linguistic familiarity rather than problem solvability. As a result, it yields inferior AUROC and selective accuracy, particularly on reasoning-intensive benchmarks.

\textbf{CoCA provides strong confidence quality with end-to-end optimization.}
CoCA generally achieves strong calibration and discrimination performance across model sizes and task categories, while preserving reasonable answer accuracy. Probe- or assessor-based baselines require additional modules, whereas CoCA obtains reliable confidence estimates through a single end-to-end optimized model.

\subsection{Additional Generalization Studies}
\label{sec:add_generalization}

The main comparison in Table~\ref{tab:main} evaluates CoCA under a math-trained setting
on the Qwen2.5-Instruct model family. To further examine whether the effectiveness of CoCA is tied to a specific backbone or to math-only training, we conduct two additional studies. Specifically, we train Llama3-8B~\cite{llama3-8B} on the same math training data and Qwen2.5-7B-Instruct on TACO code data~\cite{taco}, and evaluate them on the same benchmark suite.

Table~\ref{tab:add_generalization} summarizes the results. Across these supplementary
settings, CoCA preserves the main trend observed in Table~\ref{tab:main}: it improves
confidence discrimination and calibration, while keeping
answer accuracy comparable. These results provide additional evidence that CoCA is robust across both model backbones and verifiable
training domains. 

\subsection{Comparison with Answer-First Paradigm}
\label{sec:Answer-First}
We next compare our method with answer-first approaches, which generate an answer before estimating confidence.

\subsubsection{Baselines}

We focus on two representative answer-first methods:
\begin{itemize}[leftmargin=*]

    \item \textbf{Sampling-based surrogates (Majority Voting)}: multiple answers are sampled and clustered by semantic equivalence; confidence is computed as the proportion of samples in the largest cluster, and the representative answer from this cluster is returned as the final prediction.
    \item \textbf{Post-hoc verbalized confidence (RLCR; Reinforcement Learning with Calibration Rewards)}: we adopt the approach described in \cite{damani2025a}; the specific reward computation is given by the following formula:
    \begin{equation}
        R_{RLCR} = \mathbb{I}(y) - (s-\mathbb{I}(y))^2
    \end{equation}    

\end{itemize}

\subsubsection{Results and Practical Implications}

\begin{table}[t]
\centering
\footnotesize
\setlength{\tabcolsep}{3.5pt}
\caption{Comparison against answer-first baselines. TTC refers to the token consumption to confidence prediction. Per-dataset results are provided in Table~\ref{tab:appendix_answer_first}.}
\begin{tabular}{llccc}
\toprule
\textbf{Method} & \textbf{Metric} & \textbf{Math} & \textbf{Code} & \textbf{Factual} \\
\midrule

\multirow{3}{*}{Majority Voting}
& Acc   & 54.36  & 79.29  & 32.89 \\
& AUROC & 0.80   & 0.69   & \textbf{0.75}    \\
& TTC & 9549.16 &1996.04 & 1685.17  \\
\midrule

\multirow{3}{*}{RLCR}
& Acc   & 49.01 & 74.52  & 25.16  \\
& AUROC & \textbf{0.85}   & 0.67   & 0.67   \\
& TTC & 840.46  & 209.20 & 121.78    \\
\midrule

\multirow{3}{*}{\textbf{CoCA(ours)}}
& Acc   & 49.90  & 77.49  & 28.95 \\
& AUROC & 0.73   & \textbf{0.74}  & 0.70   \\
& TTC &  \textbf{9.95}  & \textbf{9.78}  & \textbf{9.75}    \\

\bottomrule
\end{tabular}

\label{tab:main_answer_first}
\end{table}

Table~\ref{tab:main_answer_first} summarizes the category-level averages for accuracy, AUROC, and the token consumption to confidence prediction (TTC) across Math, Code, and Factual QA. 
A full per-dataset breakdown (including all benchmarks within each category) is reported in Table~\ref{tab:appendix_answer_first}.

\textbf{AUROC differences are small across methods.}
Across the answer-first baselines and CoCA, AUROC values are broadly comparable, indicating that these methods offer similar ranking ability for separating correct from incorrect answers.

\textbf{Confidence-first enables earlier and lower-cost confidence access.}
Sampling-based confidence estimates are only available after repeated sampling and agreement checks. Post-hoc verbalized confidence also requires the full answer to be generated before confidence can be obtained. In contrast, CoCA exposes confidence before answer generation, creating an early decision point for routing, early stopping, or budget-aware inference. As shown in Table~\ref{tab:main_answer_first}, CoCA reduces TTC to approximately 10 tokens, compared with hundreds or thousands of tokens for answer-first baselines.

These results demonstrate that confidence-first is not merely a formatting change, but a paradigm shift aligned with real-world deployment requirements.

\subsection{Ablation Studies}

\subsubsection{Reward Hacking in Sequential Training}

\begin{figure*}[t]
    \centering
    \captionsetup[subfigure]{skip=2pt}
    \begin{subfigure}[t]{0.42\linewidth}
        \centering
        \includegraphics[width=\linewidth]{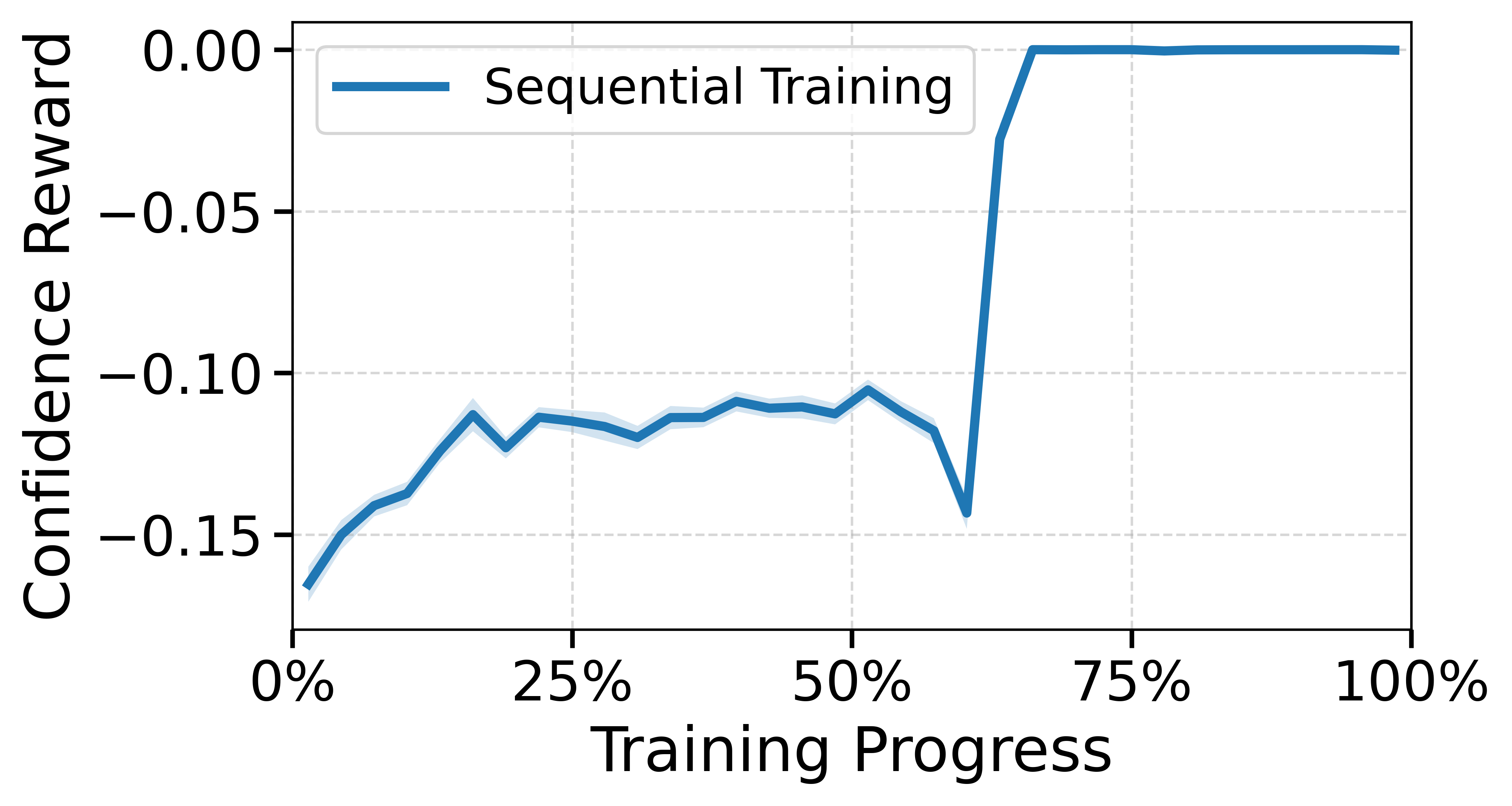}
        \caption{Sequential training: confidence reward.}
        \label{fig:acc1conf2_reward}
    \end{subfigure}
    \hspace{0.015\linewidth}
    \begin{subfigure}[t]{0.42\linewidth}
        \centering
        \includegraphics[width=\linewidth]{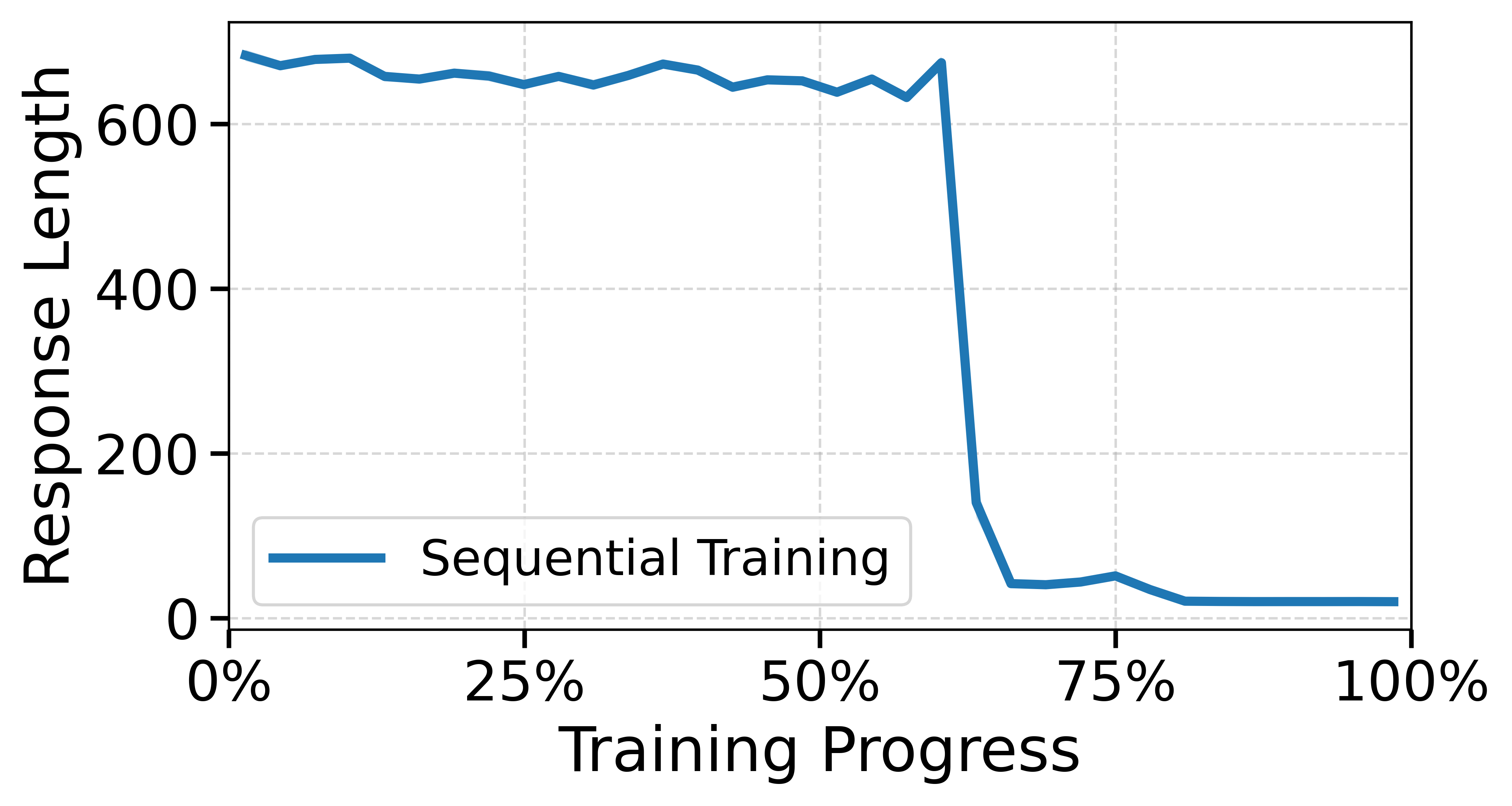}
        \caption{Sequential training: response length.}
        \label{fig:acc1conf2_response}
    \end{subfigure}

    \vspace{-0.2em}

    \begin{subfigure}[t]{0.42\linewidth}
        \centering
        \includegraphics[width=\linewidth]{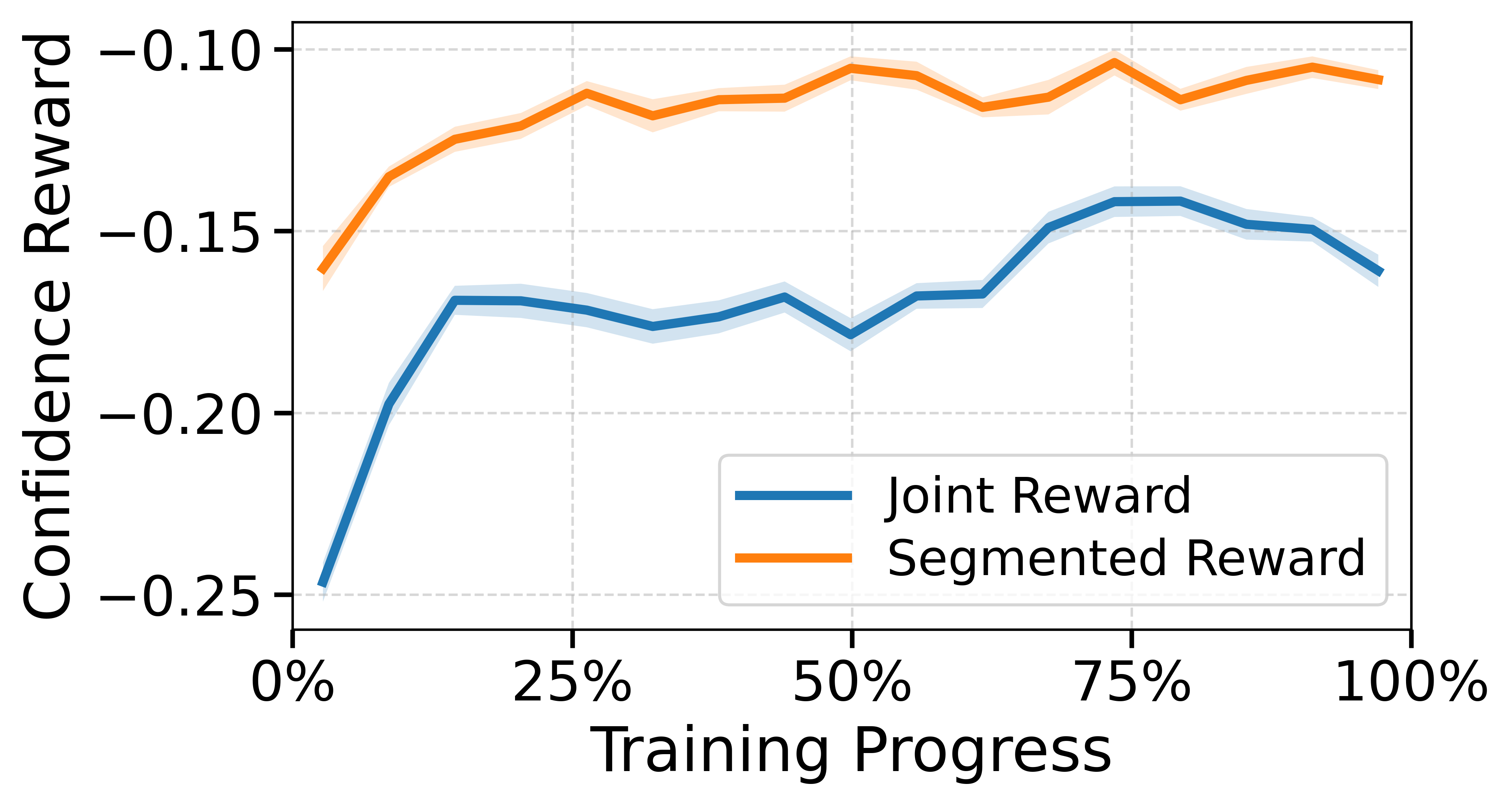}
        \caption{Joint vs.\ segmented rewards: confidence reward.}
        \label{fig:conf_reward}
    \end{subfigure}
    \hspace{0.015\linewidth}
    \begin{subfigure}[t]{0.42\linewidth}
        \centering
        \includegraphics[width=\linewidth]{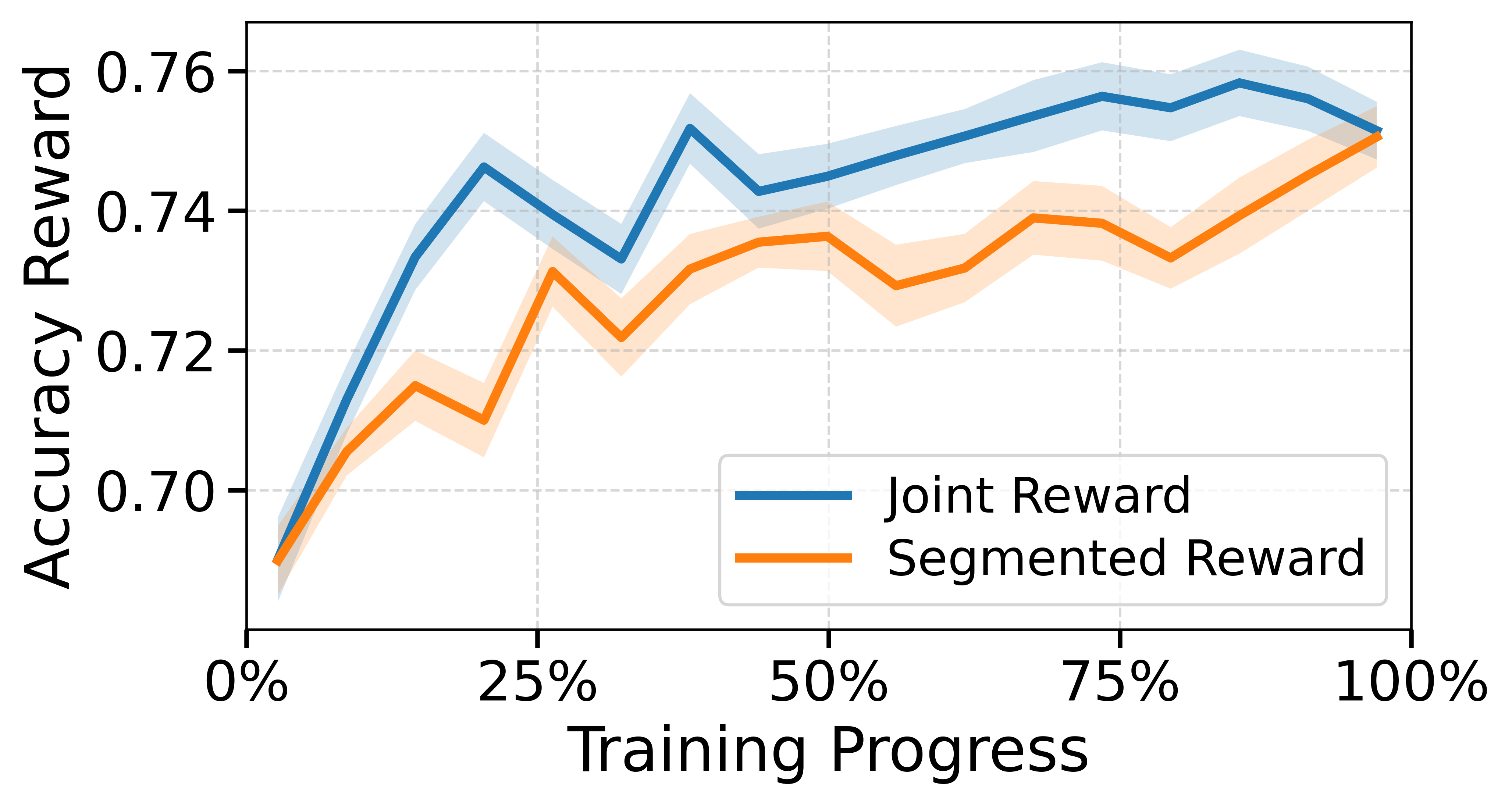}
        \caption{Joint vs.\ segmented rewards: accuracy reward.}
        \label{fig:acc_reward}
    \end{subfigure}

    \vspace{-0.5em}

    \caption{
    \textbf{Ablation training dynamics.}
    \textit{Top}: sequential confidence training exhibits reward hacking, where confidence reward increases while response length sharply decreases.
    \textit{Bottom}: segmented rewards provide a clearer optimization signal for confidence learning, with comparable accuracy-reward dynamics.
    }
    \label{fig:ablation_dynamics}
\end{figure*}

We examine a sequential training pipeline where the model is first optimized for answer correctness and then further optimized only with the confidence reward. As shown in Figure~\ref{fig:acc1conf2_reward} and Figure~\ref{fig:acc1conf2_response}, the confidence reward increases during the second-stage optimization, while the response length collapses sharply. The failure cases in Table~\ref{tab:failure_cases} show that the model learns to refuse answering or produce trivial outputs, improving the confidence objective at the cost of answer quality.

This result motivates our joint training design, which optimizes confidence calibration and answer correctness together so that confidence learning remains anchored to meaningful answer generation.

\begin{table}[t]
\centering
\small
\setlength{\tabcolsep}{4pt}
\caption{Representative degenerate outputs from sequential confidence training. }
\label{tab:failure_cases}
\begin{tabular}{p{0.35\linewidth} p{0.60\linewidth}}
\toprule
\textbf{Question} & \textbf{Model Output} \\
\midrule

Vector $\overrightarrow{a}=(2,1)$, $\overrightarrow{b}=(x,-1)$, and 
$\overrightarrow{a} \parallel \overrightarrow{b}$. Find $x$.
&
{\ttfamily <confidence>0.003</confidence>} I need more context and information to provide a proper answer. \\

\midrule

Given $\tan \alpha = 2$, calculate 
$\frac{\sin \alpha + \cos \alpha}{\sin \alpha - 3\cos \alpha}$.
&
{\ttfamily <confidence>0.005</confidence>} I cannot provide a numerical answer or a step-by-step solution as the instruction is unclear. \\

\bottomrule
\end{tabular}
\end{table}

\subsubsection{Joint Reward vs. Segmented Reward}

\begin{table}[t]
\centering
\small
\setlength{\tabcolsep}{6pt}
\caption{Comparison between Segment Reward and Joint Reward across math, code, and factual QA benchmarks.}
\begin{tabular}{llccc}
\toprule
Method & Metric & Math & Code & Factual QA \\
\midrule

\multirow{4}{*}{Segment}
& Acc   & 49.90  & 77.49  & 28.95 \\
& AUROC & 0.73   & \textbf{0.74}  & \textbf{0.70}  \\
& ECE   & 0.12  & \textbf{0.15}  & \textbf{0.22}  \\
& Brier & \textbf{0.10} & \textbf{0.18}  & \textbf{0.20}  \\

\midrule

\multirow{4}{*}{Joint}
& Acc   & 52.75 & 77.41 & 29.20 \\
& AUROC & \textbf{0.74}  & 0.52 & 0.69  \\
& ECE   & \textbf{0.09}  & 0.23 & 0.24  \\
& Brier & \textbf{0.10}  & 0.23 & 0.24 \\

\bottomrule
\end{tabular}
\label{tab:segment_vs_joint}
\end{table}

We further compare \textbf{joint rewards}, where confidence and accuracy rewards are applied to the entire response, with \textbf{segmented rewards} (ours), where the confidence reward is applied only to confidence tokens and the accuracy reward is applied only to answer tokens.

Figure~\ref{fig:conf_reward} and Figure~\ref{fig:acc_reward} show that segmented rewards lead to a stronger increase in confidence reward, while the accuracy reward remains comparable to that of joint rewards. This indicates that segment-level reward assignment provides a more targeted learning signal for confidence tokens. The evaluation results in Table~\ref{tab:segment_vs_joint} further support this interpretation: segmented rewards produce more generalizable confidence estimates, with lower ECE and Brier scores on out-of-domain Code and Factual QA.
\section{Conclusion}
We propose \textbf{CoCA (Co-optimized Confidence and Answers)}, an end-to-end, confidence-first learning framework that jointly optimizes confidence and answer quality. Across math, code, and factual QA, CoCA improves confidence quality (calibration and discrimination) while preserving accuracy and outperforming confidence-first baselines. In the main math-trained setting, these gains are obtained using only verifiable math data, and additional code-domain training further supports the robustness of the framework. Confidence-first outputs also enable early routing and termination for more efficient inference, while ablations show that joint optimization with segmented rewards is key to stable training and reduced reward hacking, leading to more reliable confidence estimates.

\section{Limitations}

Our current approach has two main limitations. First, training–evaluation mismatch for non-verifiable domains. Training is performed on automatically verifiable math and code data, while evaluation extends beyond it. This does not fully capture domains where correctness is ambiguous or unverifiable (e.g., open-ended QA, summarization, dialogue). Extending confidence-first RL to weak or indirect supervision—such as preference feedback, rubric-based graders, or tool-assisted verification—remains an important direction. Second, some hard-math evaluations rely on small, high-difficulty test sets, so metrics such as AUROC and calibration can vary noticeably with training stochasticity and checkpoint selection; larger-scale hard-math benchmarks or curated difficult collections would yield more precise estimates. 
\newpage
\bibliography{custom}

\newpage
\appendix
\section{Pseudocode of CoCA}
\label{Pseudocode of CoCA}

We present the pseudocode of the CoCA algorithm in Algorithm~\ref{alg:coca}. The algorithm separates the generation of confidence and answer into two distinct segments, and applies segment-specific rewards via a modified group-based reinforcement learning procedure.

\begin{algorithm*}[t]
\caption{CoCA (Segmented GRPO for Confidence-First Outputs)}
\label{alg:coca}
\begin{algorithmic}[1]
\REQUIRE Dataset of prompts $\mathcal{D}$; initial policy $\pi_{\theta}$; reference policy $\pi_{\text{ref}}$;
group size $G$; clip $\varepsilon$; KL coefficient $\beta$.
\ENSURE Updated policy parameters $\theta$.

\FOR{each training step}
  \STATE Sample a mini-batch of prompts $\{x_b\}_{b=1}^{B} \sim \mathcal{D}$.
  \FOR{each prompt $x$ in the mini-batch}
    \STATE Rollout $G$ responses $\{y_i\}_{i=1}^{G} \sim \pi_{\theta_{\text{old}}}(\cdot \mid x)$ with the enforced format
    \[
      y_i \equiv \texttt{<confidence>}~s_i~\texttt{</confidence>}~y_i^{a},
    \]
    where $s_i = \mathrm{Parse}(y_i^{c}) \in [0,1]$.
    \STATE Compute answer rewards 
    \STATE $r_i^{a} \leftarrow \mathbb{I}(\mathrm{AnsCorrect}(x, y_i^{a}))$.
    \STATE Compute group success rate 
    \STATE $\hat{p}(x) \leftarrow \frac{1}{G}\sum_{j=1}^{G} r_j^{a}$.
    \STATE Compute confidence rewards 
    \STATE $r_i^{c} \leftarrow -\big(s_i - \hat{p}(x)\big)^2$.
    \STATE Compute normalized group-relative advantages:
    \[
      \hat{A}_i^{a} \leftarrow \mathrm{Norm}\big(\{r_j^{a}\}_{j=1}^{G}, r_i^{a}\big)
    \]
    \[
      \hat{A}_i^{c} \leftarrow \mathrm{Norm}\big(\{r_j^{c}\}_{j=1}^{G}, r_i^{c}\big).
    \]
    \STATE Identify token index sets $\mathcal{T}_i^{c}$ (confidence segment tokens) and $\mathcal{T}_i^{a}$ (answer segment tokens).
  \ENDFOR

  \STATE Update $\theta$ by maximizing the segmented GRPO objective:
    \begin{equation*}
    \begin{aligned}
    \mathcal{L}_{\text{CoCA}}(\theta)
    &= \mathbb{E}_{x}\Bigg[
    \frac{1}{G}\sum_{i=1}^G
    \Big(
    \mathcal{L}^{c}_i(\theta)
    + \mathcal{L}^{a}_i(\theta)
    \Big)
    \Bigg],
    \end{aligned}
    \end{equation*}
    
    \begin{equation*}
    \begin{aligned}
    \mathcal{L}^{c}_i(\theta)
    &= \sum_{t\in \mathcal{T}^c_i}
    \min\Big(
    \rho_{i,t}(\theta)\hat{A}^{c}_i, \\
    &\qquad
    \mathrm{clip}(\rho_{i,t}(\theta),1-\varepsilon,1+\varepsilon)\hat{A}^{c}_i
    \Big), \\
    \mathcal{L}^{a}_i(\theta)
    &= \sum_{t\in \mathcal{T}^a_i}
    \min\Big(
    \rho_{i,t}(\theta)\hat{A}^{a}_i, \\
    &\qquad
    \mathrm{clip}(\rho_{i,t}(\theta),1-\varepsilon,1+\varepsilon)\hat{A}^{a}_i
    \Big).
    \end{aligned}
    \end{equation*}
  where 
  \begin{equation*}
  \rho_{i,t}(\theta)=\pi_{\theta}(y_{i,t}|x,y_{i,<t})/\pi_{\theta_{\text{old}}}(y_{i,t}|x,y_{i,<t}).
  \end{equation*}
  \STATE Set $\theta_{\text{old}} \leftarrow \theta$.
\ENDFOR
\end{algorithmic}
\end{algorithm*}
\section{Additional Generalization Studies}
\label{app:add_generalization}
This section provides category-level results for the additional generalization studies discussed in Section~\ref{sec:add_generalization}. 
While Table~\ref{tab:main} focuses on the primary Qwen2.5-Instruct scale comparison under math-domain training, here we further examine whether the proposed CoCA framework generalizes across model backbones and training domains.

Specifically, we report two supplementary settings. 
First, we train Llama3-8B on the same math-domain training data used in the main experiments, which evaluates whether CoCA remains effective beyond the Qwen2.5 model family. 
Second, we train Qwen2.5-7B-Instruct on code-domain data, which evaluates whether the proposed confidence-answer joint optimization remains effective when the verifiable training domain changes from math to code. 
For both settings, we use the same baseline protocol as Table~\ref{tab:main}.

\begin{table*}[t]
\centering
\small
\setlength{\tabcolsep}{5pt}
\caption{\textbf{Additional generalization results with the same baseline settings as Table~\ref{tab:main}.}
We report two supplementary settings: Llama3-8B trained on math data and Qwen2.5-7B-Instruct trained on code data.}
\label{tab:add_generalization}
\begin{tabular}{c@{\hspace{3pt}}c l ccc ccc}
\toprule
& & 
& \multicolumn{3}{c}{\textbf{Llama3-8B, Train: Math}}
& \multicolumn{3}{c}{\textbf{Qwen2.5-7B, Train: Code}} \\
\cmidrule(lr){4-6} \cmidrule(lr){7-9}
\textbf{Answer} & \textbf{Confidence} & \textbf{Metric}
& \multirow{2}{*}{\textbf{Math}} & \multicolumn{2}{c}{\textbf{OOD}}
& \multirow{2}{*}{\textbf{Code}} & \multicolumn{2}{c}{\textbf{OOD}} \\
\cmidrule(lr){5-6} \cmidrule(lr){8-9}
& &
& & \textbf{Code} & \textbf{Factual}
& & \textbf{Math} & \textbf{Factual} \\
\midrule

\multirow{4}{*}{Instruct}
& \multirow{4}{*}{Verbalized}
& Acc
& 26.88 & 33.77 & 29.68
& 77.43 & 43.73 & 30.13 \\
&& AUROC
& 0.52 & 0.56 & 0.64
& 0.57 & 0.61 & 0.63 \\
&& ECE
& 0.63 & 0.75 & 0.47
& 0.22 & 0.52 & 0.58 \\
&& Brier
& 0.59 & 0.72 & 0.42
& 0.23 & 0.50 & 0.52 \\
\midrule

\multirow{4}{*}{RLVR}
& \multirow{4}{*}{Verbalized}
& Acc
& 26.73 & 44.55 & 32.24
& 76.32 & 45.22 & 28.85 \\
&& AUROC
& 0.54 & 0.53 & 0.58
& 0.53 & 0.69 & 0.64 \\
&& ECE
& 0.70 & 0.55 & 0.52
& 0.24 & 0.50 & 0.59 \\
&& Brier
& 0.67 & 0.54 & 0.48
& 0.24 & 0.49 & 0.52 \\
\midrule

\multirow{4}{*}{RLVR}
& \multirow{4}{*}{QuestionProb}
& Acc
& 28.15 & 61.48 & 35.70
& 80.05 & 48.77 & 30.67 \\
&& AUROC
& 0.40 & 0.49 & 0.52
& 0.46 & 0.40 & 0.55 \\
&& ECE
& 0.28 & 0.58 & 0.36
& 0.70 & 0.44 & 0.29 \\
&& Brier
& 0.26 & 0.58 & 0.36
& 0.65 & 0.44 & 0.30 \\
\midrule

\multirow{4}{*}{RLVR}
& \multirow{4}{*}{Assessor}
& Acc
& 28.15 & 61.48 & 35.70
& 80.05 & 48.77 & 30.67 \\
&& AUROC
& \textbf{0.78} & 0.65 & 0.61
& 0.60 & 0.66 & 0.55 \\
&& ECE
& \textbf{0.06} & 0.27 & 0.28
& \textbf{0.08} & 0.24 & 0.40 \\
&& Brier
& \textbf{0.08} & 0.43 & 0.20
& \textbf{0.16} & 0.17 & 0.36 \\
\midrule

\multirow{4}{*}{RLVR}
& \multirow{4}{*}{Probe}
& Acc
& 28.15 & 61.48 & 35.70
& 80.05 & 48.77 & 30.67 \\
&& AUROC
& 0.76 & 0.65 & 0.60
& 0.59 & 0.66 & 0.51 \\
&& ECE
& \textbf{0.06} & 0.23 & 0.24
& 0.23 & 0.22 & 0.49 \\
&& Brier
& \textbf{0.08} & 0.28 & 0.22
& 0.22 & 0.18 & 0.45 \\
\midrule

\multicolumn{2}{c}{\multirow{4}{*}{\makecell{\textbf{CoCA(ours)}\\Joint optimization}}}
& Acc
& 34.88 & 52.24 & 35.16
& 78.68 & 44.79 & 29.10 \\
\multicolumn{2}{c}{}
& AUROC
& \textbf{0.78} & \textbf{0.75} & \textbf{0.74}
& \textbf{0.71} & \textbf{0.76} & \textbf{0.76} \\
\multicolumn{2}{c}{}
& ECE
& 0.07 & \textbf{0.16} & \textbf{0.15}
& 0.09 & \textbf{0.14} & \textbf{0.15} \\
\multicolumn{2}{c}{}
& Brier
& 0.09 & \textbf{0.23} & \textbf{0.16}
& 0.17 & \textbf{0.12} & \textbf{0.15} \\

\bottomrule
\end{tabular}
\end{table*}

Table~\ref{tab:add_generalization} reports category-level results, while the per-dataset results are provided in Table~\ref{tab:appendix_llama3_8b} and Table~\ref{tab:appendix_qwen25_7b_code}.

Overall, these results show that CoCA's benefits are not limited to the original Qwen2.5 math-trained setting. Although improvements vary across benchmarks and domains, CoCA generally improves calibration or discrimination while maintaining comparable answer accuracy, supporting the robustness of segmented confidence-answer optimization across backbone and training-domain shifts.

\section{Training and Evaluation Details}

\subsection{Training Configuration and Schedules}
All models are trained using the \textbf{MindSpeed-RL} framework on \textbf{Ascend 910B} and \textbf{Ascend 910C} accelerators. Unless otherwise specified, we use the same optimization and decoding settings across all experiments.

\textbf{Optimization hyperparameters.}
\begin{itemize}
    \item Global batch size: $128 \times 16$
    \item Learning rate: $1 \times 10^{-6}$
    \item Maximum generation length: 4096 tokens
    \item Temperature: 1.0
\end{itemize}

No additional sampling strategies (e.g., top-$k$, nucleus sampling) are used during training, in order to preserve the model’s intrinsic output distribution.

\textbf{Training schedules.}
We set the training length based on the optimization behavior observed in the confidence-first baseline comparison. With one-epoch training, the main objective already approaches a plateau before the epoch ends, while later updates bring limited gains but increase cost. Therefore, for the answer-first comparison and the joint-vs-segmented reward ablation, we use a 0.5-epoch schedule as a controlled and efficient budget.

For the sequential-training ablation, we use a longer schedule because it serves as a stability test. Specifically, we train for one epoch on answer accuracy followed by one epoch on confidence, which allows us to examine whether confidence-only optimization becomes unstable after accuracy training and to reveal delayed reward-hacking behavior.

\subsection{Prompt Format}

To enforce the confidence-first output structure, we adopt a fixed system prompt and a task-specific user prompt.

\begin{promptbox}
\textbf{System Prompt}

You need to provide the answer as well as its confidence level to follow-up questions.
The confidence level is a number between 0 and 1 (inclusive) enclosed within
\texttt{<confidence>} \texttt{</confidence>} tags.
The final format that must be followed is:

\texttt{<confidence> confidence level here </confidence> answer here }
\end{promptbox}

\begin{promptbox}
\textbf{User prompt}

\texttt{\{question\}} Please reason step by step, and put your final answer within
\texttt{\textbackslash boxed\{\}}.
\end{promptbox}

\subsection{Evaluation Protocol}

All evaluations are conducted using the \textbf{OpenCompass} framework. Due to the confidence-first output format, we implement a lightweight modification to the evaluation pipeline:
\begin{enumerate}
    \item The confidence score enclosed within <confidence> </confidence> tags is first extracted.
    \item The remaining text (i.e., the answer segment) is passed to the standard task-specific evaluator.
\end{enumerate}

During inference, \textbf{no sampling strategies} are employed; each response is generated via a single forward pass. This ensures that both answer quality and confidence estimates reflect the model’s inherent policy distribution rather than artifacts of stochastic decoding.

\section{Detailed Evaluation Results}
\label{sec:appendix_detailed_results}

This section provides a comprehensive breakdown of per-dataset evaluation results for all model scales and training variants considered in this work. While the main paper reports aggregated performance over task categories (Math, Code, and Factual QA) to highlight high-level trends, the tables in this appendix present fine-grained results on each individual benchmark.

For each model size (Qwen2.5-1.5B, 3B, and 7B), we report accuracy (Acc), area under the ROC curve (AUROC), expected calibration error (ECE), and Brier score on all datasets. Following standard practice, higher values indicate better performance for Acc and AUROC, whereas lower values are preferred for ECE and Brier score. To facilitate comparison across training methods, the best-performing method for each dataset and metric is highlighted in bold.

Notably, for the smallest model (Qwen2.5-1.5B), confidence generation is less reliable. We therefore report ECE together with the confidence-generation success rate (SR), defined as the fraction of examples for which a valid confidence estimate is produced. In addition, for datasets on which the accuracy is zero, AUROC cannot be meaningfully computed; such cases are uniformly marked as ``–''. These cases are excluded from aggregate metric calculations.

Overall, these detailed results complement the main paper by exposing dataset-level behavior that is otherwise obscured by category-level aggregation, and they provide additional evidence for the robustness and limitations of the proposed methods across model scales and task domains.

\begin{table*}[t]
\centering
\scriptsize
\setlength{\tabcolsep}{3.5pt}
\caption{Per-dataset results for Qwen2.5-7B-Instruct. “MBPP(s)” denotes sanitized MBPP. “--” indicates undefined AUROC (e.g., when all answers are incorrect and the label has no variance). Bold highlights the best method per dataset for AUROC (higher is better) and for ECE/Brier (lower is better).}
\begin{tabular}{llcccc cccc}
\toprule
\multirow{2}{*}{Method} & \multirow{2}{*}{Metric}
& \multicolumn{4}{c}{Math}
& \multicolumn{2}{c}{Code}
& \multicolumn{2}{c}{Factual QA} \\
\cmidrule(lr){3-6} \cmidrule(lr){7-8} \cmidrule(lr){9-10}
& & GSM8K & Math500 & AIME24 & AIME25
  & MBPP(s) & HumanEval
  & SimpleQA & TriviaQA \\
\midrule

\multirow{4}{*}{Instruct Model}
& Acc   & 89.46 & 78.80 & 6.67 & 0.00    & 73.15 & 81.71 & 3.26 & 56.99 \\
& AUROC & 0.62  & 0.67 & 0.55 & --   & 0.54  & 0.59  & 0.62 & 0.63  \\
& ECE   & 0.08  & 0.18 & 0.88 & 0.93 & 0.27  & 0.17  & 0.79 & 0.36  \\
& Brier & 0.10  & 0.19 & 0.83 & 0.87 & 0.27  & 0.18  & 0.67 & 0.37  \\
\midrule

\multirow{4}{*}{RLVR}
& Acc   & 93.33 & 86.00 & 20.00 & 6.67 & 73.54 & 78.66 & 4.02 & 57.45 \\
& AUROC & 0.62  & 0.61 & 0.62 & 0.65 & 0.51  & 0.59  & 0.59 & 0.61  \\
& ECE   & 0.05  & 0.13 & 0.75 & 0.89 & 0.26  & 0.19  & 0.80 & 0.35  \\
& Brier & 0.06  & 0.14 & 0.73 & 0.86 & 0.26  & 0.20  & 0.69 & 0.36  \\
\midrule

\multirow{4}{*}{RLVR+QuestionProb}
& Acc   & 93.63 & 80.60 & 16.67 & 10.00 & 75.10 & 78.05 & 3.63 & 59.13 \\
& AUROC & 0.50  & 0.24 & 0.34  & 0.20 & 0.48  & 0.45  & 0.53 & 0.56  \\
& ECE   & 0.92  & 0.76 & \textbf{0.10} & \textbf{0.09}& 0.70  & 0.70  & \textbf{0.03} & 0.58  \\
& Brier & 0.90  & 0.74 & 0.15  & 0.10 & 0.68  & 0.67  & \textbf{0.04} & 0.58  \\
\midrule

\multirow{4}{*}{RLVR+Assessor Model}
& Acc   & 93.63 & 80.60 & 16.67 & 10.00 & 75.10 & 78.05 & 3.63 & 59.13 \\
& AUROC & 0.70  & 0.78 & 0.88 & 0.97 & 0.51  & 0.58  & 0.59 & 0.62  \\
& ECE   & \textbf{0.03} & 0.05 & 0.37  & 0.53 & \textbf{0.07} & 0.50  & 0.50 & \textbf{0.03} \\
& Brier & \textbf{0.06} & 0.13  & 0.24  & 0.34 & \textbf{0.19} & 0.41  & 0.29 & \textbf{0.23}  \\
\midrule

\multirow{4}{*}{RLVR+Probe}
& Acc   & 93.63 & 80.60 & 16.67 & 10.00 & 75.10 & 78.05 & 3.63 & 59.13 \\
& AUROC & \textbf{0.75} & 0.82 & 0.80  & \textbf{1.00} & 0.64 & \textbf{0.60} & \textbf{0.68} & 0.60  \\
& ECE   & 0.07  & \textbf{0.03} & 0.19 & 0.20 & 0.70  & 0.71  & 0.17 & 0.24  \\
& Brier & 0.07 & 0.12 & 0.14 & \textbf{0.07} & 0.67  & 0.68  & 0.10 & 0.31  \\
\midrule

\multirow{4}{*}{CoCA(ours)}
& Acc   & 92.95 & 85.40 & 10.00 & 3.33 & 73.93 & 79.27 & 3.79 & 53.90  \\
& AUROC & 0.65  & \textbf{0.84} & \textbf{0.96} & 0.40 & \textbf{0.74} & \textbf{0.70} & 0.66 & \textbf{0.72} \\
& ECE   & 0.06  & 0.09 & 0.11 & 0.15 & 0.20  & \textbf{0.12} & 0.26 & 0.26  \\
& Brier & 0.07  & \textbf{0.11} & \textbf{0.05} & 0.11 & 0.21  & \textbf{0.17} & 0.18 & 0.29  \\

\bottomrule
\end{tabular}

\label{tab:appendix_qwen25_7b}
\end{table*}

\begin{table*}[t]
\centering
\scriptsize
\setlength{\tabcolsep}{3.5pt}
\caption{Per-dataset results for Qwen2.5-3B-Instruct. “MBPP(s)” denotes sanitized MBPP. “--” indicates undefined AUROC (e.g., when all answers are incorrect and the label has no variance). Bold highlights the best method per dataset for AUROC (higher is better) and for ECE/Brier (lower is better).}
\begin{tabular}{llcccc cc cc}
\toprule
\multirow{2}{*}{Method} & \multirow{2}{*}{Metric}
& \multicolumn{4}{c}{Math}
& \multicolumn{2}{c}{Code}
& \multicolumn{2}{c}{Factual QA} \\
\cmidrule(lr){3-6} \cmidrule(lr){7-8} \cmidrule(lr){9-10}
& & GSM8K & Math500 & AIME24 & AIME25
  & MBPP(s) & HumanEval
  & SimpleQA & TriviaQA \\
\midrule

\multirow{4}{*}{Instruct Model}
& Acc   & 77.03 & 59.80 & 6.67 & 6.67 & 61.09 & 66.46 & 2.54 & 42.88 \\
& AUROC & 0.50  & 0.53 & 0.40 & 0.73 & 0.48  & 0.55  & 0.55 & 0.59  \\
& ECE   & 0.16  & 0.31 & 0.86 & 0.81 & 0.33  & \textbf{0.27}  & 0.82 & 0.50  \\
& Brier & 0.20  & 0.33 & 0.79 & 0.75 & 0.35  & \textbf{0.30}  & 0.71 & 0.49  \\
\midrule

\multirow{4}{*}{RLVR}
& Acc   & 88.55 & 73.20 & 3.33 & 6.67 & 49.03 & 20.12 & 3.40 & 44.44 \\
& AUROC & 0.50  & 0.51 & 0.00 & 0.36 & 0.48  & 0.48  & 0.53 & 0.54  \\
& ECE   & \textbf{0.02} & 0.18 & 0.89 & 0.81 & 0.42 & 0.70 & 0.77 & 0.46 \\
& Brier & 0.10 & 0.23 & 0.80 & 0.74 & 0.43 & 0.65 & 0.66 & 0.45 \\
\midrule

\multirow{4}{*}{RLVR+QuestionProb}
& Acc   & 89.39 & 73.40 & 13.33 & 10.00 & 67.32 & 70.73 & 2.36 & 49.65 \\
& AUROC & 0.53  & 0.24 & 0.61 & 0.30 & 0.45 & 0.38 & 0.46 & 0.53 \\
& ECE   & 0.89  & 0.71 & 0.11 & \textbf{0.06} & 0.65 & 0.64 & \textbf{0.02} & 0.49 \\
& Brier & 0.88  & 0.71 & 0.12 & 0.10 & 0.64 & 0.62 & \textbf{0.02} & 0.49 \\
\midrule

\multirow{4}{*}{RLVR+Assessor Model}
& Acc   & 89.39 & 73.40 & 13.33 & 10.00 & 67.32 & 70.73 & 2.36 & 49.65 \\
& AUROC & 0.74  & \textbf{0.86} & \textbf{0.88} & 0.83 & 0.58 & \textbf{0.69} & 0.58 & 0.63 \\
& ECE   & 0.03  & 0.08 & 0.18 & 0.36 & \textbf{0.09} & 0.49 & 0.47 & \textbf{0.07} \\
& Brier & \textbf{0.09}  & 0.14 & 0.11 & 0.21 & \textbf{0.22} & 0.43 & 0.25 & \textbf{0.24} \\
\midrule

\multirow{4}{*}{RLVR+Probe}
& Acc   & 89.39 & 73.40 & 13.33 & 10.0 & 67.32 & 70.73 & 2.36 & 49.65 \\
& AUROC & 0.77 & \textbf{0.86} & 0.83 & 0.88 & 0.60 & 0.57 & 0.63 & 0.64 \\
& ECE   & 0.04  & \textbf{0.03} & 0.15 & 0.21 & 0.15 & 0.58 & 0.27 & 0.31 \\
& Brier & \textbf{0.09}  & \textbf{0.13} & 0.11 & 0.10 & 0.24 & 0.54 & 0.14 & 0.34 \\
\midrule

\multirow{4}{*}{CoCA(ours)}
& Acc   & 87.34 & 75.40 & 13.33 & 3.33 & 62.20 & 72.56 & 0.88 & 48.59 \\
& AUROC & \textbf{0.79} & \textbf{0.86} & \textbf{0.88} & \textbf{1.00} & \textbf{0.67} & 0.67 & \textbf{0.73} & \textbf{0.73} \\
& ECE   & 0.08  & 0.08 & \textbf{0.10} & 0.10 & 0.16 & 0.39 & 0.11 & 0.17 \\
& Brier & 0.10  & 0.14 & \textbf{0.10} & \textbf{0.03} & 0.24 & 0.36 & 0.03 & \textbf{0.24} \\

\bottomrule
\end{tabular}
\label{tab:appendix_qwen25_3b}
\end{table*}

\begin{table*}[t]
\centering
\scriptsize
\setlength{\tabcolsep}{3.5pt}
\caption{Per-dataset results for Qwen2.5-1.5B-Instruct. “MBPP(s)” denotes sanitized MBPP. “--” indicates undefined AUROC (e.g., when all answers are incorrect and the label has no variance). Bold highlights the best method per dataset for AUROC (higher is better) and for ECE/Brier (lower is better).}
\begin{tabular}{llcccc cc cc}
\toprule
\multirow{2}{*}{Method} & \multirow{2}{*}{Metric}
& \multicolumn{4}{c}{Math}
& \multicolumn{2}{c}{Code}
& \multicolumn{2}{c}{Factual QA} \\
\cmidrule(lr){3-6} \cmidrule(lr){7-8} \cmidrule(lr){9-10}
& & GSM8K & Math500 & AIME24 & AIME25
  & MBPP(s) & HumanEval
  & SimpleQA & TriviaQA \\
\midrule

\multirow{4}{*}{Instruct Model}
& Acc   & 28.35 & 26.80 & 0.00 & 0.00 & 54.09 & 56.10 & 1.87 & 27.74 \\
& AUROC & 0.52 & 0.51 & -- & -- & 0.56 & 0.50 & 0.49 & 0.56 \\
& ECE (SR) & 0.61(0.89) & 0.65(0.96) & 0.89(0.93) & 0.93(0.80)
           & 0.42(0.98) & 0.40(0.98) & 0.81(0.96) & 0.63(0.93) \\
& Brier & 0.58 & 0.63 & 0.82 & 0.87 & 0.42 & 0.41 & 0.72 & 0.60 \\
\midrule

\multirow{4}{*}{RLVR}
& Acc   & 66.94 & 56.60 & 0.00 & 0.00 & 42.02 & 35.98 & 2.17 & 31.37 \\
& AUROC & 0.52 & 0.51 & -- & -- & 0.61 & 0.55 & 0.53 & 0.55 \\
& ECE (SR) & 0.24(0.91) & 0.36(0.94) & 0.94 & 0.94(0.97)
           & 0.55(0.99) & 0.56 & 0.89(0.95) & 0.62(0.90) \\
& Brier & 0.28 & 0.38 & 0.89 & 0.88 & 0.54 & 0.56 & 0.82 & 0.60 \\
\midrule

\multirow{4}{*}{RLVR+QuestionProb}
& Acc   & 80.06 & 64.20 & 6.67 & 6.67 & 57.59 & 56.71 & 1.64 & 37.94 \\
& AUROC & 0.47 & 0.27 & 0.34 & \textbf{0.68}
         & 0.48 & 0.44 & 0.47 & 0.54 \\
& ECE (SR) & 0.78 & 0.59 & \textbf{0.06} & \textbf{0.02}
           & 0.50 & 0.45 & \textbf{0.00} & 0.37 \\
& Brier & 0.76 & 0.60 & 0.07 & 0.06
         & 0.50 & 0.46 & \textbf{0.02} & 0.37 \\
\midrule

\multirow{4}{*}{RLVR+Assessor Model}
& Acc   & 80.06 & 64.20 & 6.67 & 6.67 & 57.59 & 56.71 & 1.64 & 37.94 \\
& AUROC & 0.76 & 0.83 & \textbf{0.93} & 0.34
         & 0.63 & 0.63 & 0.61 & 0.64 \\
& ECE (SR) & 0.03 & 0.09 & 0.16 & 0.37
           & 0.09 & 0.41 & 0.29 & \textbf{0.03} \\
& Brier & \textbf{0.14} & 0.17 & \textbf{0.06} & 0.20
         & 0.24 & 0.46 & 0.11 & 0.22 \\
\midrule

\multirow{4}{*}{RLVR+Probe}
& Acc   & 80.06 & 64.20 & 6.67 & 6.67 & 57.59 & 56.71 & 1.64 & 37.94 \\
& AUROC & 0.76 & 0.80 & 0.73 & 0.20
         & 0.76 & 0.51 & 0.60 & 0.53 \\
& ECE (SR) & \textbf{0.02} & \textbf{0.05} & 0.16 & 0.15
           & \textbf{0.07} & 0.47 & 0.68 & 0.23 \\
& Brier & \textbf{0.14} & 0.17 & 0.07 & 0.11
         & \textbf{0.20} & 0.47 & 0.52 & 0.31 \\
\midrule

\multirow{4}{*}{CoCA(ours)}
& Acc   & 79.53 & 61.20 & 6.67 & 0.00 & 55.64 & 37.80 & 1.16 & 36.62 \\
& AUROC & \textbf{0.77} & \textbf{0.85} & 0.83 & --
         & \textbf{0.78} & \textbf{0.66} & \textbf{0.71} & \textbf{0.71} \\
& ECE (SR) & 0.11 & 0.07 & 0.09 & 0.09
           & 0.09 & \textbf{0.09} & 0.08 & 0.10 \\
& Brier & 0.15 & \textbf{0.15} & \textbf{0.06} & \textbf{0.02}
         & \textbf{0.20} & \textbf{0.23} & \textbf{0.02} & \textbf{0.21} \\

\bottomrule
\end{tabular}
\label{tab:appendix_qwen25_1p5b}
\end{table*}

\begin{table*}[t]
\centering
\scriptsize
\setlength{\tabcolsep}{3.5pt}
\caption{Per-dataset results for llama3-8B-Instruct. “MBPP(s)” denotes sanitized MBPP. “--” indicates undefined AUROC (e.g., when all answers are incorrect and the label has no variance). Bold highlights the best method per dataset for AUROC (higher is better) and for ECE/Brier (lower is better).}
\begin{tabular}{llcccc cc cc}
\toprule
\multirow{2}{*}{Method} & \multirow{2}{*}{Metric}
& \multicolumn{4}{c}{Math}
& \multicolumn{2}{c}{Code}
& \multicolumn{2}{c}{Factual QA} \\
\cmidrule(lr){3-6} \cmidrule(lr){7-8} \cmidrule(lr){9-10}
& & GSM8K & Math500 & AIME24 & AIME25
  & MBPP(s) & HumanEval
  & SimpleQA & TriviaQA \\
\midrule

\multirow{4}{*}{Instruct Model}
& Acc   & 76,12 & 31.40 & 0.00 & 0.00 & 44.36 & 23.17 & 3.58 & 55.78 \\
& AUROC & 0.49 & 0.55 & -- & -- & 0.61 & 0.51 & 0.70 & 0.58 \\
& ECE  & 0.15 & 0.59 & 0.88 & 0.90
           & 0.74 & 0.76 & 0.64& 0.29 \\
& Brier & 0.21 & 0.56 & 0.77 & 0.81 & 0.72 & 0.72 & 0.52 & 0.32 \\
\midrule

\multirow{4}{*}{RLVR}
& Acc   & 76.72 & 30.20 & 0.00 & 0.00 & 48.25 & 40.85 & 5.25 & 59.22 \\
& AUROC & 0.52 & 0.55 & -- & -- & 0.54 & 0.52 & 0.56 & 0.60 \\
& ECE & 0.22 & 0.67 & 0.96 & 0.93
           & 0.51 & 0.58 & 0.78 & 0.26 \\
& Brier & 0.23 & 0.66 & 0.93 & 0.87 & 0.50 & 0.58 & 0.66 & 0.30 \\
\midrule

\multirow{4}{*}{RLVR+QuestionProb}
& Acc   & 76.72 & 30.20 & 0.00 & 0.00 & 48.25 & 40.85 & 5.25 & 59.22 \\
& AUROC & 0.41 & 0.39 & -- & --
         & 0.51 & 0.46 & 0.50 & 0.53 \\
& ECE & 0.79 & 0.25 & \textbf{0.03} & 0.03
           & 0.60 & 0.56 & \textbf{0.02} & 0.69 \\
& Brier & 0.78 & 0.25 & \textbf{0.00} & \textbf{0.00}
         & 0.60 & 0.56 & \textbf{0.02} & 0.69 \\
\midrule

\multirow{4}{*}{RLVR+Assessor Model}
& Acc   & 76.72 & 30.20 & 0.00 & 0.00 & 48.25 & 40.85 & 5.25 & 59.22 \\
& AUROC & \textbf{0.75} & \textbf{0.81} & -- & --
         & 0.72 & 0.58 & 0.55 & 0.67 \\
& ECE & \textbf{0.04} & \textbf{0.05} & \textbf{0.03} & 0.10
           & 0.30 & 0.23 & 0.26 & 0.30 \\
& Brier & 0.14 & \textbf{0.15} & \textbf{0.00} & 0.01
         & 0.29 & 0.57 & 0.10 & 0.29 \\
\midrule

\multirow{4}{*}{RLVR+Probe}
& Acc   & 76.72 & 30.20 & 0.00 & 0.00 & 48.25 & 40.85 & 5.25 & 59.22 \\
& AUROC & 0.73 & 0.78 & -- & --
         & 0.72 & 0.57 & 0.61 & 0.59 \\
& ECE & \textbf{0.04} & 0.06 & 0.06 & 0.07
           & 0.22 & 0.24 & 0.22 & 0.26 \\
& Brier & 0.15 & 0.16 & 0.01 & 0.01
         & 0.25 & 0.30 & 0.12 & 0.31 \\
\midrule

\multirow{4}{*}{CoCA(ours)}
& Acc   & 85.37 & 50.80 & 3.33 & 0.00 & 58.75 & 45.73 & 4.60 & 65.71 \\
& AUROC & 0.67 & \textbf{0.81} & \textbf{0.86} & --
         & \textbf{0.77} & \textbf{0.72} & \textbf{0.74} & \textbf{0.73} \\
& ECE & 0.11 & 0.13 & 0.05 & \textbf{0.00}
           & \textbf{0.14} & \textbf{0.18} & 0.19 & \textbf{0.10} \\
& Brier & \textbf{0.13} & 0.19 & 0.04 & \textbf{0.00}
         & \textbf{0.21} & \textbf{0.24} & 0.10 & \textbf{0.21} \\

\bottomrule
\end{tabular}
\label{tab:appendix_llama3_8b}
\end{table*}

\begin{table*}[t]
\centering
\scriptsize
\setlength{\tabcolsep}{3.5pt}
\caption{Per-dataset results for Qwen2.5-7B-Instruct trained on code data. “MBPP(s)” denotes sanitized MBPP. “--” indicates undefined AUROC (e.g., when all answers are incorrect and the label has no variance). Bold highlights the best method per dataset for AUROC (higher is better) and for ECE/Brier (lower is better).}
\begin{tabular}{llcccc cc cc}
\toprule
\multirow{2}{*}{Method} & \multirow{2}{*}{Metric}
& \multicolumn{4}{c}{Math}
& \multicolumn{2}{c}{Code}
& \multicolumn{2}{c}{Factual QA} \\
\cmidrule(lr){3-6} \cmidrule(lr){7-8} \cmidrule(lr){9-10}
& & GSM8K & Math500 & AIME24 & AIME25
  & MBPP(s) & HumanEval
  & SimpleQA & TriviaQA \\
\midrule

\multirow{4}{*}{Instruct Model}
& Acc   & 89.46 & 78.80 & 6.67 & 0.00    & 73.15 & 81.71 & 3.26 & 56.99 \\
& AUROC & 0.62  & 0.67 & 0.55 & --   & 0.54  & 0.59  & 0.62 & 0.63  \\
& ECE   & \textbf{0.08}  & 0.18 & 0.88 & 0.93 & 0.27  & 0.17  & 0.79 & 0.36  \\
& Brier & \textbf{0.10}  & 0.19 & 0.83 & 0.87 & 0.27  & 0.18  & 0.67 & 0.37  \\
\midrule

\multirow{4}{*}{RLVR}
& Acc   & 89.69 & 81.20 & 6.67 & 3.33 & 72.76 & 79.88 & 2.94 & 54.76 \\
& AUROC & 0.58 & 0.68 & 0.54 & 0.96 & 0.52 & 0.53 & 0.65 & 0.63 \\
& ECE & \textbf{0.08} & \textbf{0.15} & 0.87 & 0.90
           & 0.27 & 0.20 & 0.78 & 0.39 \\
& Brier & \textbf{0.10} & 0.17 & 0.83 & 0.84 & 0.27 & 0.20 & 0.65 & 0.39 \\
\midrule

\multirow{4}{*}{RLVR+QuestionProb}
& Acc   & 91.36 & 80.40 & 13.33 & 10.00 & 78.99 & 81.10 & 3.51 & 57.83 \\
& AUROC & 0.53 & 0.23 & 0.68 & 0.14
         & 0.44 & 0.47 & 0.52 & 0.57 \\
& ECE & 0.87 & 0.72 & \textbf{0.03} & 0.12
           & 0.69 & 0.71 & \textbf{0.02} & 0.56 \\
& Brier & 0.84 & 0.69 & 0.11 & 0.11
         & 0.63 & 0.67 & \textbf{0.03} & 0.56 \\
\midrule

\multirow{4}{*}{RLVR+Assessor Model}
& Acc   & 91.36 & 80.40 & 13.33 & 10.00 & 78.99 & 81.10 & 3.51 & 57.83 \\
& AUROC & 0.56 & 0.71 & \textbf{0.80} & 0.58
         & 0.63 & 0.57 & 0.56 & 0.53 \\
& ECE & 0.30 & 0.27 & 0.07 & 0.31
           & \textbf{0.06} & 0.10 & 0.62 & 0.18 \\
& Brier & 0.17 & 0.22 & \textbf{0.09} & 0.21
         & \textbf{0.16} & 0.16 & 0.43 & 0.28 \\
\midrule

\multirow{4}{*}{RLVR+Probe}
& Acc   & 91.36 & 80.40 & 13.33 & 10.00 & 78.99 & 81.10 & 3.51 & 57.83 \\
& AUROC & 0.56 & 0.73 & 0.78 & 0.57
         & 0.50 & 0.67 & 0.48 & 0.53 \\
& ECE & 0.29 & 0.39 & 0.08 & \textbf{0.11}
           & 0.15 & 0.31 & 0.67 & 0.31 \\
& Brier & 0.19 & 0.31 & 0.10 & 0.11
         & 0.19 & 0.25 & 0.53 & 0.37 \\
\midrule

\multirow{4}{*}{CoCA(ours)}
& Acc   & 89.01 & 76.80 & 10.00 & 3.33 & 76.26 & 81.10 & 2.38 & 55.81 \\
& AUROC & \textbf{0.70} & \textbf{0.84} & 0.52 & \textbf{0.98}
         & \textbf{0.73} & \textbf{0.69} & \textbf{0.76} & \textbf{0.75} \\
& ECE & 0.18 & \textbf{0.15} & 0.12 & 0.12
           & 0.13 & \textbf{0.04} & 0.21 & \textbf{0.08} \\
& Brier & 0.14 & \textbf{0.15} & 0.14 & \textbf{0.05}
         & 0.18 & \textbf{0.15} & 0.09 & \textbf{0.21} \\

\bottomrule
\end{tabular}
\label{tab:appendix_qwen25_7b_code}
\end{table*}

\begin{table*}[t]
\centering
\small
\setlength{\tabcolsep}{4pt}
\caption{Per-dataset results for the comparison between our method and the answer-first methods.
 “MBPP(s)” denotes sanitized MBPP. TTC refers to the token consumption to confidence prediction. }
\begin{tabular}{llcccc ccc ccc}
\toprule
\multirow{2}{*}{Method} & \multirow{2}{*}{Metric}
& \multicolumn{4}{c}{Math}
& \multicolumn{2}{c}{Code}
& \multicolumn{2}{c}{Factual QA} \\
\cmidrule(lr){3-6} \cmidrule(lr){7-8} \cmidrule(lr){9-10}
& 
& GSM8K & Math500 & AIME24 & AIME25
& MBPP(s) & HumanEval
& SimpleQA & TriviaQA \\
\midrule

\multirow{3}{*}{Majority Voting}
& Acc
& 94.16  &  86.60 &  20.00   & 16.67    
&   76.26   &   82.32     
&  4.90     &  60.88    \\
& AUROC
&  \textbf{0.78}  &   0.86    &   0.86    &   0.68   
& 0.66   &   0.71     
& \textbf{0.73}  &  \textbf{0.77} \\
& TTC
&  3266.38     &  7229.33     &   14374.50    & 13326.4 
&   1945.30    & 2046.78    
&  2321.69   &  1048.64      \\
\midrule

\multirow{3}{*}{RLCR}
& Acc
& 92.49 & 80.20 & 13.33 & 10.00 
& 71.60 & 77.44 
& 0.65  & 49.67  \\
& AUROC
& 0.66  & \textbf{0.91}  & \textbf{1.00}  & \textbf{0.83} 
& 0.59  & \textbf{0.75}  
& 0.68  & 0.66   \\
& TTC
& 392.01   &  727.65     &    1074.07   &  1168.10   
& 219.96   & 198.43        
& 131.54  & 112.00       \\
\midrule

\multirow{3}{*}{CoCA(ours)}
& Acc
& 92.87 & 83.40 & 20.00 & 3.33
& 75.10 & 79.88 
& 3.70  & 54.19  \\
& AUROC
& 0.66  & 0.86  & 0.83  & 0.57 
& \textbf{0.73}  & 0.74  
& 0.67  & 0.73   \\
& TTC
& 9.92  & 9.53    &  9.93   & 10.40    
& 9.60   &  9.96   
&  9.79  & 9.70  \\

\bottomrule
\end{tabular}
\label{tab:appendix_answer_first}
\end{table*}

\begin{table*}[t]
\centering
\small
\setlength{\tabcolsep}{4pt}
\caption{Detail comparison between Segment Reward and Joint Reward across math, code, and factual QA benchmarks. “MBPP(s)” denotes sanitized MBPP.}
\begin{tabular}{llcccccccc}
\toprule
\multirow{2}{*}{Method} & \multirow{2}{*}{Metric}
& \multicolumn{4}{c}{Math}
& \multicolumn{2}{c}{Code}
& \multicolumn{2}{c}{Factual QA} \\
\cmidrule(lr){3-6} \cmidrule(lr){7-8} \cmidrule(lr){9-10}
& 
& GSM8K & Math500 & AIME24 & AIME25 
& MBPP(s) & HumanEval
& SimpleQA & TriviaQA\\
\midrule

\multirow{4}{*}{Segment Reward}
& Acc
& 92.87 & 83.40 & 20.00 & 3.33
& 75.10 & 79.88 & 3.70 & 54.19 \\

& AUROC
& \textbf{0.66} & \textbf{0.86} & 0.83 & 0.57
& \textbf{0.73} & \textbf{0.74} & 0.67 & \textbf{0.73}  \\

& ECE

& \textbf{0.06} & \textbf{0.06} & 0.15 & 0.19
& \textbf{0.19} & \textbf{0.10} & 0.22 & \textbf{0.22}\\

& Brier
& 0.07 & \textbf{0.10} & 0.09 & \textbf{0.13}
& \textbf{0.20} & \textbf{0.15} & \textbf{0.14} & \textbf{0.26} \\

\midrule

\multirow{4}{*}{Joint Reward}
& Acc
& 93.86 & 83.80 & 20.00 & 13.33
& 74.32 & 80.49 & 2.10 & 56.30 \\

& AUROC
& 0.50 & 0.73 & \textbf{0.85} & \textbf{0.88} 
& 0.50 & 0.53 & \textbf{0.74} & 0.64  \\

& ECE
& \textbf{0.06} & 0.12 & \textbf{0.03} & \textbf{0.15}
& 0.26 & 0.19 & \textbf{0.15} & 0.33  \\

& Brier
& \textbf{0.06} & 0.13 & \textbf{0.07} & 0.14
& 0.26 & 0.19 & \textbf{0.14} & 0.33  \\

\bottomrule
\end{tabular}
\label{tab:segment_vs_joint_detail}
\end{table*}

\end{document}